# Understanding why shooters shoot – An AI-powered engine for basketball performance profiling

Alejandro Rodriguez Pascual | UC San Diego
Ishan Mehta | UC San Diego
Muhammad Khan | UC San Diego
Frank Rodriz | UC San Diego
Rose Yu | UC San Diego

## 1. Introduction

In professional basketball, strategy is everything. It is crucial for the coaching staff of a team to analyze an opposing team and develop an effective strategy. Understanding player shooting profiles is an essential part of this analysis: knowing where certain opposing players like to shoot from can help coaches neutralize offensive gameplans from their opponents. Understanding where their players are most comfortable can lead them to developing more effective offensive strategies.

However, existing approaches for collecting this information are time-consuming: watching hours upon hours of film. Doing so just to identify player performance profiles takes time away from developing strategies, practicing those strategies to nail the execution, or can even cut into coaches' free time. Thus, an automatic tool that can provide these performance profiles in a timely manner can become invaluable for coaches to maximize both the effectiveness of their game plan as well as the time dedicated to practice and other related activities.

Additionally, basketball is dictated by many variables, such as playstyle and game dynamics, that can change the flow of the game and, by extension, player performance profiles. It is crucial that the performance profiles can reflect the diverse playstyles, as well as the fast-changing dynamics of the game. A tool that can deliver interpretable representations of player performance profiles while taking these other variables into account would be the ideal approach.

## 2. Previous work

Previously, we developed a Multiresolution Tensor Learning (MRTL) [1] algorithm that models the complex relations between ball-handler, defenders and their positions on the court as latent factors. Additionally, we expanded this MRTL [1] algorithm into a Spatiotemporal Multiresolution Tensor Learning (ST-MRTL) [2] algorithm that incorporates non-spatial dimensions, such as time, into the learning process.

## 2.1. Tensor Learning

Our goal is to generate interpretable performance profiles that can predict whether a given ball handler will shoot within the next second, given his position on the court and the relative positions of the defenders around him. The input $X$ is the one-hot encoding of the court positions. $W$ is the parameter tensor of the model. $Y_i \in \{0, 1\}$ is the binary output equal to 1 if player $i$ shoots within the next second, and $\sigma$ is the sigmoid function. We instantiate a tensor classification model as follows:

$$Y_i = \sum_{d^1=1}^{D_r^1} \sum_{d^2=1}^{D_r^2} \sigma\left(W_{i,d^1,d^2}^{(r)} X_{i,d^1,d^2}^{(r)}\right) \quad (1)$$

where $i \in \{1,\ldots,I\}$ is the ball-handler ID and $r$ is the spatial resolution. Here $d^1$ indexes the ballhandler's position on the discretized court of dimension $D_r^1$, and $d^2$ indexes the relative defender positions around the ball-handler in a discretized grid of dimension $D_r^2$. We assume that only defenders close to the ball-handler affect shooting probability and set $D_r^2 < D_r^1$, to reduce dimensionality.

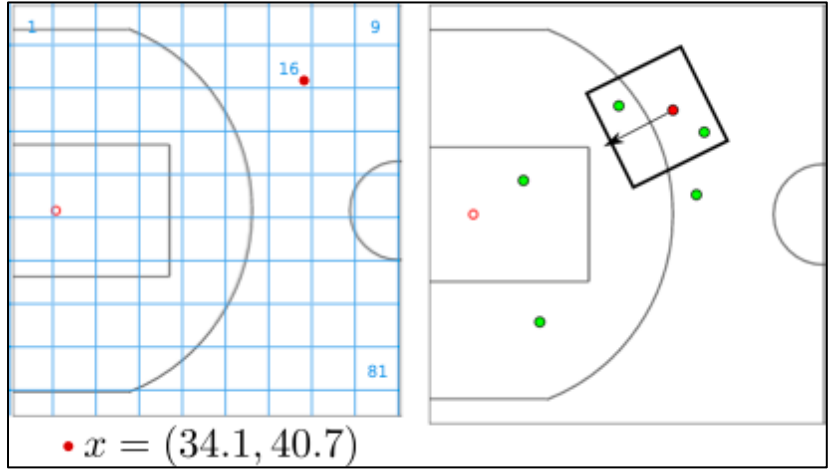

**Figure 1**. Left: Discretizing a continuous-valued position of a player (red) via a spatial grid. Right: sample frame with a ball-handler (red) and defenders (green). Only defenders close to the ball-handler are considered.

As shown in Figure 1, given the ballhandler's position on the court and the relative positions of the defenders around him, we orient the defender positions so that the direction from the ball-handler to the basket points up. For the ball handler location in $D_r^1$, we discretize the half-court into resolutions $4 \times 5, 8 \times 10, 20 \times 25, 40 \times 50$. For the relative defender locations, at the full resolution, we choose a $12 \times 12$ grid around the ball handler where the ball handler is located at $(6, 2)$ (more space in front of the ball handler than behind them). We also consider a smaller grid around the ball handler for the defender locations, assuming that defenders that are far away from the ball handler do not influence shooting probability. We use $6 \times 6$, $12 \times 12$ for defender positions.

## 2.2. Multiresolution Tensor Learning (MRTL)

To reduce the number of the parameters in $W$, we assume a low-dimensional latent structure in $W$ which can characterize distinct patterns in the data and also alleviate model overfitting. We assume this structure by decomposing $W$ into latent factors [4][5]. The low-rank tensor learning model is as follows:

$$Y_i = \sum_{d^1=1}^{D_r^1} \sum_{d^2=1}^{D_r^2} \sum_{k=1}^{K} \sigma\left(A_{i,k}^{(r)} B_{d^1,k}^{(r)} C_{d^2,k}^{(r)} X_{i,d^1,d^2}^{(r)}\right) \qquad (2)$$

Here $K$ is a hyper-parameter that controls the number of distinct performance profiles. Matrix $A$ represents the individual player's style. $B$ and $C$ are the common ball-handler and defender spatial profiles across all players.

We first train an approximate full-rank tensor model in (1) at a low resolution. The weight tensor is then decomposed into latent factors and these values are used to initialize the low-rank model. The low-rank model in (2) is then trained to the final desired accuracy. As we use approximately optimal solutions of the full-rank model as initializations for the low-rank model, our algorithm produces interpretable latent factors in a variety of different scenarios and datasets.

In general, learning a high-dimensional tensor model is computationally expensive and memory inefficient. We utilize multiple resolutions for this issue. We train both the full rank and low rank models at multiple resolutions, starting from a coarse spatial resolution and progressively increasing the resolution. At each resolution $r$, we learn $W^{(r)}$ using stochastic gradient descent. When the stopping criterion is met, we transform $W^{(r)}$ to $W^{(r+1)}$ in a process we call finegraining [1]. Once the full-rank resolution has been trained up to resolution $r_0$ (which can be chosen to fit GPU memory or time constraints), we decompose $W^{(r)}$ using tensor factorization. Then the low-rank model is trained at different resolutions to final desired accuracy, finegraining [1] to move to the next resolution.

In particular, the basketball data contains two spatial modes: the ball handler's position and the relative defender positions around the ball handler. Let us denote the pair of resolutions as $(D_r^1, D_r^2)$. We train the full-rank model at resolutions $(4 \times 5, 6 \times 6)$, $(8 \times 10, 6 \times 6)$, $(8 \times 10, 12 \times 12)$ and the low-rank model at resolutions $(8 \times 10, 12 \times 12)$, $(20 \times 25, 12 \times 12)$, $(40 \times 50, 12 \times 12)$.

## 2.3. Spatiotemporal Multiresolution Tensor Learning (ST-MRTL)

The ST-MRTL [2] algorithm is very similar to the MRTL [1] algorithm described in 2.2 above, with the main difference being the addition of a *non-spatial* dimension to the tensor model learning process. We instantiate a tensor classification model as follows:

$$Y_i = \sum_{t=1}^{T_s} \sum_{d^1=1}^{D_r^1} \sum_{d^2=1}^{D_r^2} \sigma\left(W_{i,t,d^1,d^2}^{(r,s)} X_{i,t,d^1,d^2}^{(r,s)}\right) \qquad (3)$$

where $i \in \{1, \ldots, I\}$ is the ball-handler ID, $r$ is the spatial resolution and $s$ is the temporal non-spatial resolution. Here $t$ indexes the value of the non-spatial dimension with respect to the entire frame, $d^1$ indexes the ballhandler's position on the discretized court of dimension $D_r^1$, and $d^2$ indexes the relative defender positions around the ball-handler in a discretized grid of dimension $D_r^2$. Similarly, the low-rank tensor learning model is as follows:

$$Y_i = \sum_{t=1}^{T_s} \sum_{d^1=1}^{D_r^1} \sum_{d^2=1}^{D_r^2} \sum_{k=1}^{K} \sigma\left(A_{i,k}^{(r,s)} B_{t,k}^{(r,s)} C_{d^1,k}^{(r,s)} D_{d^2,k}^{(r,s)} X_{i,t,d^1,d^2}^{(r,s)}\right) \quad (4)$$

Here the matrix $B$ represents the global factor for either the time or playstyle. $C$ and $D$ are the common ball-handler and defender spatial profiles. The model training process is the same as described above in 2.1 and 2.2, with the only difference being that not only do we finegrain [1] on a spatial resolution $r$, but we can also do so on a non-spatial resolution $s$ (e.g.: we can change the discretization of the data on a temporal dimension from one entire game to four different quarters).

## 3. Methods

We can use an MRTL [1] model to generate interpretable heatmaps that represent player performance profiles. However, shooting profiles vary greatly depending on a player's playstyle and game dynamics, so using an MRTL [1] model alone would fail to capture these factors. We instead resort to an extension of MRTL. We developed two approaches:

1. Applying a ST-MRTL [2] model with a feature such as time or playstyle as a non-spatial dimension: this method allows the model to factor out the influence a non-spatial dimension might have on player performance profiles, resulting in clearer representations of them. Non-spatial factors are still saved to a tensor, which can be used to weight and re-order the visualizations by order of importance for each value of the non-spatial dimension.

2. Creating a Dynamic MRTL model that extends the spatial discretization to differentiate between spatial data across a non-spatial dimension: this method allows the model to generate heatmaps of player performance profiles specific to each non-spatial feature, and we can even implement regularization between features when appropriate. Non-spatial factors are intrinsic to the visualizations of performance profiles, so the only way to analyze the influence of the non-spatial factors would be to compare all the visualizations.

Additionally, from (2) and (4), we can obtain weights from $A$ that identify which heatmaps define each player's performance profile more and less. We can also consider the weights from $B$ in (4) to also include whatever effect a non-spatial factor may have. As such, we can use these weights to weigh and reorder the performance profiles represented by the heatmaps, which allows us to not analyze the performance profiles of specific players, but also how different playstyles affect player performance profiles or how they might change throughout the game.

### 3.1. Data Preprocessing for Playstyle and Game Dynamics

To account for the effect that the time of the game may have on shooting profiles, we included the quarter number of each case in the dataset. This allows us to differentiate between quarters and analyze how a player's performance profile evolves throughout the game.

To define the playstyle of each player, we tried to separate players into 7 distinct clusters. This was done using a dataset available on the NBA's website[1]. We used synergy playstyle frequencies and points per possession to assign each player to a given playstyle based on how often and how effective they were at a given play-type. We wanted to use information that was not available in the coordinate location set, so we could give our model new information.

Synergy play-type data is separated into 11 different play-types: Isolation, Transition, Pick and Roll Ball Handler, Pick and Roll Roller, Spot Up, Cuts, Post-ups, Putbacks, Off-Screen and Miscellaneous. These tell us how a player got the shooting look they attempted rather than where they got it.

We created a table based on each player's frequency and efficiency in each play-type along with their total volume. This was a 23-dimensional data set. In order to avoid dimensionality issues, we applied a PCA transformation to the dataset and then used the three most important principal components. This allowed us to both visualize the data, and better fit a model to it.

We then applied K-Means clustering to create 7 cluster centers (based on prior knowledge of NBA player roles and silhouette score plots). Thus, each player was assigned to one of: Ball-Handling Guards, Catch & Shoot Guards, Perimeter Wings, Versatile Wings, Stretch 4s, Rolling Bigs and Post-Up Bigs.

### 3.2. Applying the Spatiotemporal Multiresolution Tensor Learning (ST-MRTL) Model

We can use an ST-MRTL [2] model with time or playstyle as a non-spatial dimension. Not only would this add a dimension to $W$, as shown in (3), but it would also capture these variables in their own latent factor ($B$ in (4)). This would allow the model to factor non-spatial variables, such as time or playstyle, independently from spatial variables, which could yield clearer player performance profiles while also allowing us to examine how these factors influence them.

When factoring how time affects shooting profiles, we used the quarter number as the non-spatial (temporal) feature. We also finegrained [1] from one full game to 4 quarters during the training process. This means the model starts making no distinction between quarters, and then separates them later on. While we didn't implement temporal regularization for this approach, it may have a positive effect.

When factoring how playstyle affects shooting profiles, we used the players' play-style cluster as the non-spatial feature. We also finegrained [1] from one group to 7 groups during the learning process. This means the model starts making no distinction between playstyles, and then separates them later on. Because playstyles can be quite different, and have no correlation between them, we implemented no forms of regularization between them.

---

[1] NBA's website: www.nba.com

### 3.3. Using a Dynamic Multiresolution Tensor Learning (MRTL) Model

We can extend the spatial discretization of the data to differentiate between time periods or playstyles. The training process is the same as the MRTL [1] algorithm, only with a bigger model. The discretization method is defined by the following:

$$x = x_0 + f \cdot x_{max}^{(r)} \tag{5}$$

for vector $x$, where vector $x_0$ is the initial $x$ from $D_r^1$, vector $f$ is the set of values of the non-spatial feature (where 0 is the first value), and $x_{max}^{(r)}$ is a constant for the size of the $x$-dimension of $D_r^1$, defined by the spatial resolution $r$. This method discretizes our spatial data in a way that essentially allows us to build multiple courts side-to-side where players and defenders could fall into.

When factoring how time affects shooting profiles, we use the scheme described in (5) to create 4 adjacent courts, one per quarter. This means that two game plays in different quarters where the ball-handler is in the same position would fall into different courts, allowing the model to differentiate between quarters. Additionally, we implemented temporal regularization [3] between quarters with a regularization coefficient $\lambda$ of 0.0002. This results in performance profiles looking similar between quarters, allowing for the creation of dynamic player performance profiles that evolve throughout the course of a game.

When factoring how playstyle affects shooting profiles, we use the scheme described in (5) to create 7 adjacent courts, one per playstyle. This means that two game situations from players with different playstyles where the ball-handler is in the same position would fall into different courts, allowing the model to differentiate between playstyles. Because playstyle can be quite different, and have no correlation between them, we implemented no forms of regularization between them.

## 4. Results & Discussion

We used a dataset of all games in the 2015-2016 NBA season[2] collected by Second Spectrum to test our methods. This allows us to determine how effective our approach is at identifying playstyle-specific and dynamic shooting patterns, regardless of offensive system or team.

We compare the test F1 score of each approach, training the model with an adaptive learning rate, training on a maximum of 50 epochs per resolution, and a stop condition of validation loss increase.

### 4.1. Temporal Dimension

The measured test F1 scores of both of our approaches, as well as those of the MRTL [1] model, are shown in Table 1. Both methods performed similarly to the MRTL [1] model. The dynamic MRTL model performed the best, but it took significantly longer to train than the ST-MRTL [2] model.

---

[2] Dataset can be accessed here: https://github.com/sealneaward/nba-movement-data

**Table 1**

Test F1 Scores (Temporal factor, defined by 4 quarters)

|  | Base MRTL Model | Dynamic MRTL Model | ST-MRTL Model |
| --- | --- | --- | --- |
| Test F1 Score | 0.66 | 0.72 | 0.68 |

We also visualize the player performance profiles from the Dynamic MRTL and ST-MRTL [2] models as heatmaps. The Dynamic MRTL model generates heatmaps specifically for each quarter, then weighs and reorders them for each player. The ST-MRTL [2] model weighs and reorders a shared set of general heatmaps for each quarter and player.

We selected Russell Westbrook as an example to test the effectiveness of our methods at visualizing player performance profiles. The heatmaps of Russell Westbrook for each quarter generated by the Dynamic MRTL model are shown in Figure 2, and those generated by the ST-MRTL [2] model are shown in Figure 3.

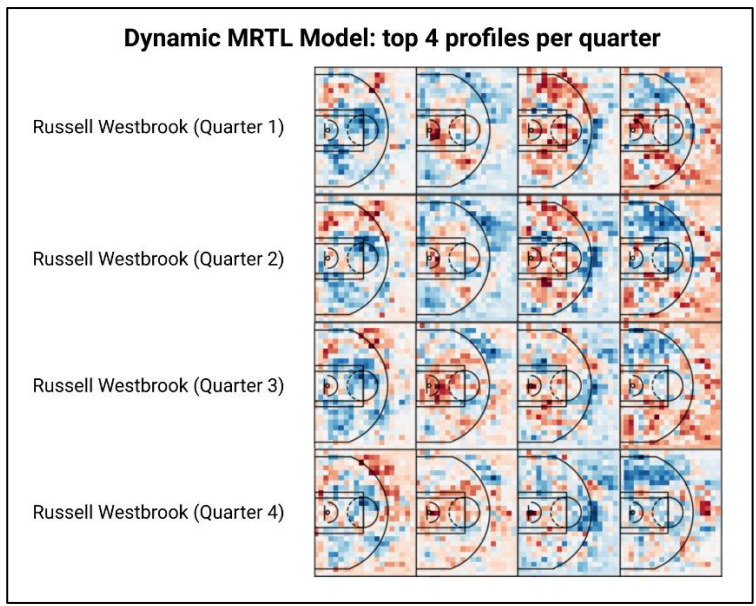

**Figure 2:** top 4 heatmaps for each quarter for Russell Westbrook, generated by the Dynamic MRTL model.

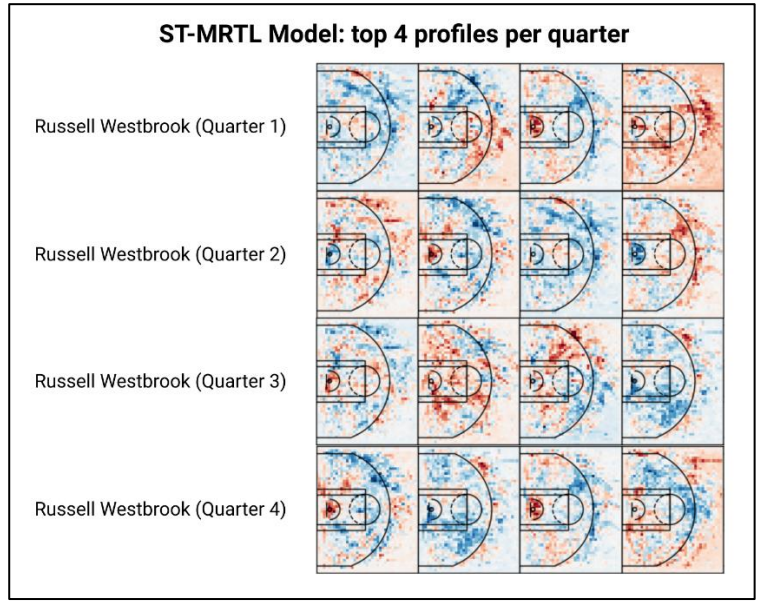

**Figure 3:** top 4 heatmaps for each quarter for Russell Westbrook, generated by the ST-MRTL [2] model.

Both methods yielded interpretable heatmaps. For example, the heatmaps generated by the Dynamic MRTL model show that Westbrook shoots more three-pointers as the game progresses, since the defense forces him to take those shots. This is shown by the growing red spots on the right-wing three-pointer area in the first heatmap for each quarter, as well as the third heatmap having the inside of the three-point line get bluer as the game goes on. It also shows that he will still attack the basket throughout the entire game, as the second heatmap for each quarter has a red spot near the basket, albeit diminishing as the game goes on. The heatmaps generated by the ST-MRTL [2] model tell a similar story, as every quarter has at least 1 heatmap with a red area under the basket, and other heatmaps show that he likes shooting three-pointers from the right wing.

### 4.2. Playstyle Dimension

The measured test F1 scores of both of our approaches on both datasets, as well as those of the base MRTL [1] model, are shown in Table 2. Both of our methods performed similarly to the MRTL [1] model. The dynamic MRTL model performed the best, but it took significantly longer to train than the ST-MRTL [2] model.

**Table 2**

Test F1 Scores (Playstyle factor, defined by 7 playstyle clusters)

|  | Base MRTL Model | Modified MRTL Model | ST-MRTL Model |
| --- | --- | --- | --- |
| Test F1 Score | 0.66 | 0.73 | 0.68 |

We also visualize the player performance profiles from the Dynamic MRTL and ST-MRTL [2] models as heatmaps. The Dynamic MRTL model generates heatmaps specifically for each playstyle, then weighs and reorders them for each player. The ST-MRTL [2] model weighs and reorders a shared set of general heatmaps for each playstyle and player.

We selected Russell Westbrook and Kevin Love, as they are two players with noticeably different playstyles (a Ball-Handling Guard and a Stretch 4). The heatmaps of Russell Westbrook and Kevin Love, as well as their playstyles, generated by the Dynamic MRTL model are shown in Figure 4, and those generated by the ST-MRTL [2] model are shown in Figure 5.

Both methods yielded interpretable heatmaps. For example, the heatmaps generated by the Dynamic MRTL model show that Ball-Handling Guards like the mid-range area, as well as corner 3s and three-pointers from the top of the key and the right wing. Stretch 4s, on the other hand, have a much bigger emphasis on corner 3s and a few mid-range areas around the elbows and the sides. We also get good visualizations of the profiles of both of our players: Westbrook has a big emphasis on three-pointers on the right wing and top of the key, as well as more specific mid-range spots on the wings (which likely correspond to his frequently used pull-up jumper). Love, on the other hand, has a bigger emphasis on the three-point line and under the basket (likely since he's a good rebounder), and near the elbows (probably corresponding to him taking one dribble inside the three-point line whenever defenders would contest his three-point shot). The heatmaps generated by the ST-MRTL [2] model tell a similar story, although they seem to better capture Westbrook's tendency to attack the basket and Love's tendency to shoot three-pointers from the wings. Overall, it seems the ST-MRTL [2] model generates clearer, more informative heatmaps than the Dynamic MRTL model.

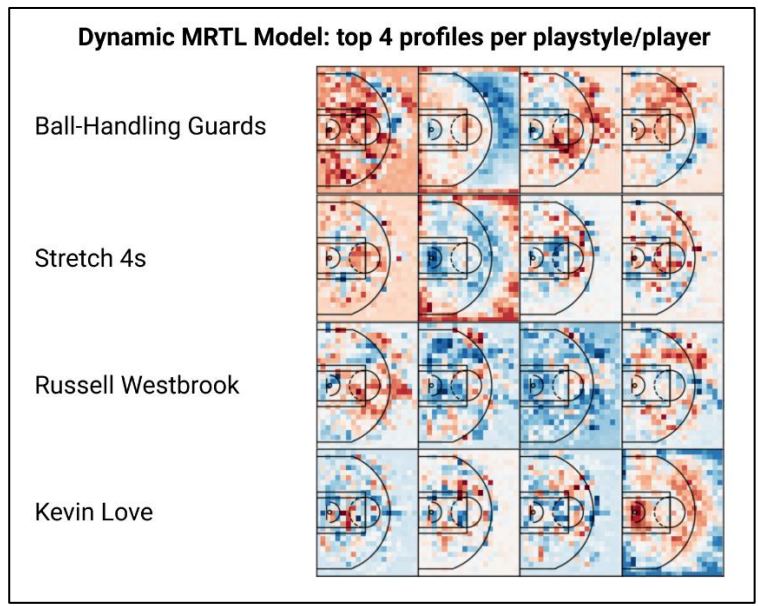

**Figure 4:** top 4 heatmaps for each for Russell Westbrook and Kevin Love and their playstyles, generated by the Dynamic MRTL model.

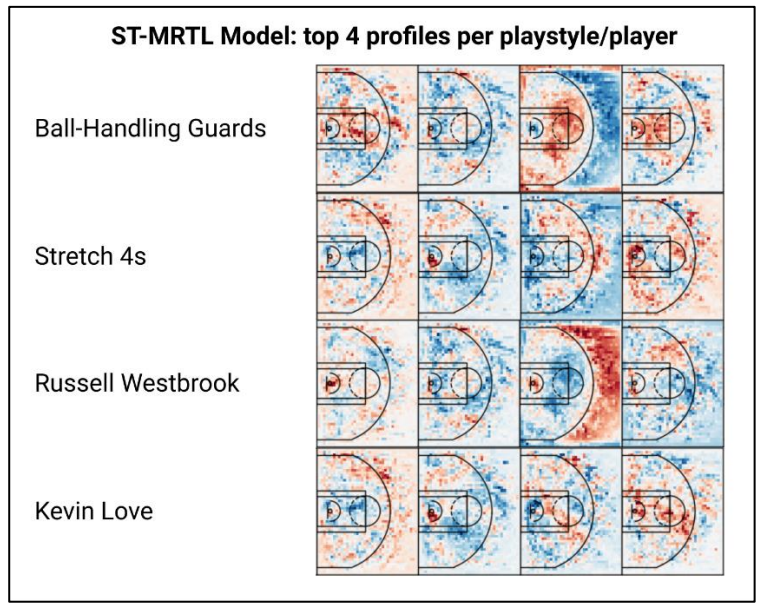

**Figure 5**: top 4 heatmaps for Russell Westbrook and Kevin Love and their playstyles, generated by the ST-MRTL [2] model.

### 4.3. General Heatmaps

Both the ST-MRTL [2] model and the Dynamic MRTL model generate heatmaps for specific players and specific quarters/playstyles. However, only the ST-MRTL [2] is also able to generate general heatmaps that represent the average behavior of all players captured in our dataset. These heatmaps serve the same purpose as the heatmaps generated by the MRTL [1] model: to visualize and understand the average behavior of all players captured in our dataset. We compare some general heatmaps from the MRTL [1] model and the ST-MRTL [2] model in Figure 6.

Both the MRTL [1] and ST-MRTL [2] generate interpretable visualizations of performance profiles. However, it seems the ST-MRTL [2] model does a better job at creating a distinction between "hot" and "cold" areas. This leads to a sharper contrast in the heatmaps between red and blue areas, which makes the heatmaps clearer.

When using a temporal dimension, it seems the heatmaps are less specific than those generated when using a playstyle dimension. For example, the heatmaps on the playstyle dimension have very specific "hot" zones in the right corner and on the left side of the basket, as well as three pointers from the top of the key. On the other hand, those generated on the temporal dimension have "hot" areas around the basket and around the three-point line, with few cases were a specific spot stands out. Overall, the ST-MRTL [2] model generates clearer general heatmaps than the MRTL [1] model, with playstyle dimension providing the most precise visualizations of performance profiles.

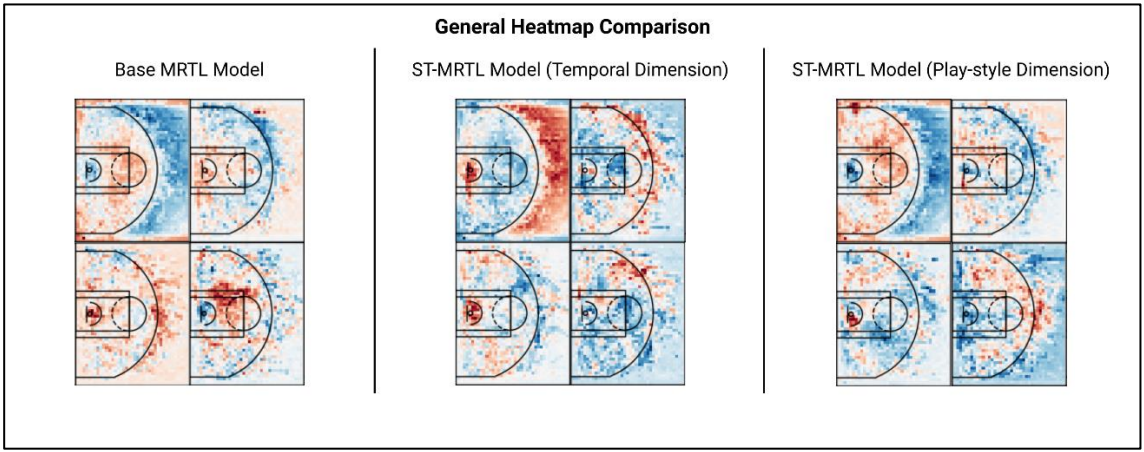

**Figure 6:** comparison of 4 general heatmaps from the MRTL [1] and ST-MRTL [2] models.

# 5. Conclusion & Future Work

We present a tool that can visualize player performance profiles in a timely manner while taking into account factors such as play-style and game dynamics. Our approach generates interpretable heatmaps that allow us to identify and analyze how non-spatial factors, such as game dynamics or playstyle, affect player performance profiles. Our methods perform significantly better than a base MRTL [1] model and provide clearer visualizations of player performance profiles with and without these non-spatial factors. Overall, our methods provide an effective and efficient tool that can provide insight into how certain players and teams play, without requiring the time-consuming process of reviewing hours of film. Our tool can be particularly useful for coaches and basketball fans around the world. We believe our tool can also be applied to other similar sports, such as baseball or soccer, albeit with different datasets and results.

Future work in this area could involve applying some sort of regularization across the non-spatial dimension in the ST-MRTL [2] model, improving the MRTL [1] or ST-MRTL [2] algorithms to improve effectiveness or efficiency, or improving the finegraining [1] or tensor decomposition [4][5] processes for increased accuracy or efficiency.

# References

[1] Park, Jung Yeon, et al. "Multiresolution Tensor Learning for Efficient and Interpretable Spatial Analysis." *International Conference on Machine Learning*. ICML, 2020.
[2] Walker and Yu. "Multiresolution Tensor Learning for Efficient and Interpretable Spatiotemporal Analysis." *AI for Earth Sciences Workshop* at NeurIPS, 2020.
[3] Yu, Hsiang-Fu, Nikhil Rao, and Inderjit S. Dhillon. "Temporal regularized matrix factorization for high-dimensional time series prediction." *Advances in neural information processing systems* at NeurIPS, 2016: 847-855.
[4] Hitchcock, F. L. "The expression of a tensor or a polyadic as a sum of products." *Journal of Mathematics and Physics*, 6(1-4):164–189, 1927.
[5] Kolda, T. G. and Bader, B. W. "Tensor Decompositions and Applications." *SIAM Review*, 51(3):455–500, 2009. ISSN 0036-1445. doi: 10.1137/07070111X.